\documentclass{article}

\usepackage{amssymb,amsmath}
\usepackage{graphicx}
\usepackage{wrapfig}
\usepackage{verbatim}
\usepackage{pstricks}

\def \1 { {\mathbf{1}} }
\def \I { {\mathbf{I}} }

\newcommand{\mb}{\mathbf}

\def \1 { {\mathbf{1}} }
\def \I { {\mathbf{I}} }

\title{
The Natural Gradient by Analogy to Signal Whitening, and Recipes and Tricks for its Use
}
\author{Jascha Sohl-Dickstein\\
Redwood Center for Theoretical Neuroscience\\
University of California at Berkeley
}
\begin{document}

\maketitle

The natural gradient, as introduced by \cite{Amari1987}, allows for more efficient gradient descent by removing dependencies and biases inherent in a function's parameterization.  Several papers present the topic thoroughly and precisely \cite{Amari1987,Amari1998,Amari2000,Theis2005,Amari2010}.  It remains a very difficult idea to get your head around however.  The intent of this note is to provide simple intuition for the natural gradient and its uses.  
We review how an ill conditioned parameter space can undermine learning, introduce the natural gradient by analogy to the more widely understood concept of signal whitening, and present tricks and specific prescriptions for applying the natural gradient to learning problems.  
To our knowledge, this is the first time a connection has been made between signal whitening and the natural gradient.  

\section{Natural gradient}

\subsection{A simple example \label{simple example}}

We begin with a simple probabilistic model which has clearly been very poorly parametrized.  For this we use a two dimensional gaussian distribution, with means written in terms of the parameters $\theta \in \mathcal R^2$,
\begin{eqnarray}
q\left( \mb x; \theta \right) = \frac{1}{2 \pi}\exp\left[ -\frac{1}{2}\left( x_1 - \left[3 \theta_1 + \frac{1}{3} \theta_2 \right]\right)^2 - \frac{1}{2}\left( x_2 - \left[\frac{1}{3} \theta_1 \right]\right)^2 \right]
.
\end{eqnarray}
As an objective function $J\left( \theta \right)$ we use the negative log likelihood of $q\left( \mb x; \theta \right)$ under an observed data distribution $p\left( \mb x \right)$
\begin{eqnarray}
\label{log like obj}
J\left( \theta \right) = 
-\left< \log q\left( \mb x; \theta \right) \right>_{p\left( \mb x \right)}
.
\end{eqnarray}
Using steepest gradient descent to minimize the negative log likelihood involves taking steps like
\begin{eqnarray}
\Delta \theta &  \propto & -\nabla_\theta J\left( \theta \right) \\
\label{descent equation}
\left[
\begin{matrix}
\Delta \theta_1 \\
\Delta \theta_2
\end{matrix}
\right]
 & \propto & 
\left[
\begin{matrix}
	\left<
		3 \left( x_1 - \left[3 \theta_1 + \frac{1}{3} \theta_2 \right]\right)
		+
		\frac{1}{3} \left( x_2 - \left[\frac{1}{3} \theta_1 \right]\right)
	\right>_{p\left( \mb x \right)} \\
	\left<
		\frac{1}{3} \left( x_1 - \left[3 \theta_1 + \frac{1}{3} \theta_2 \right]\right)
	\right>_{p\left( \mb x \right)}
\end{matrix}
\right]
.
\end{eqnarray}

\begin{figure}
\begin{center}
\begin{tabular}{cc}
(a)\includegraphics[width=0.4\linewidth]{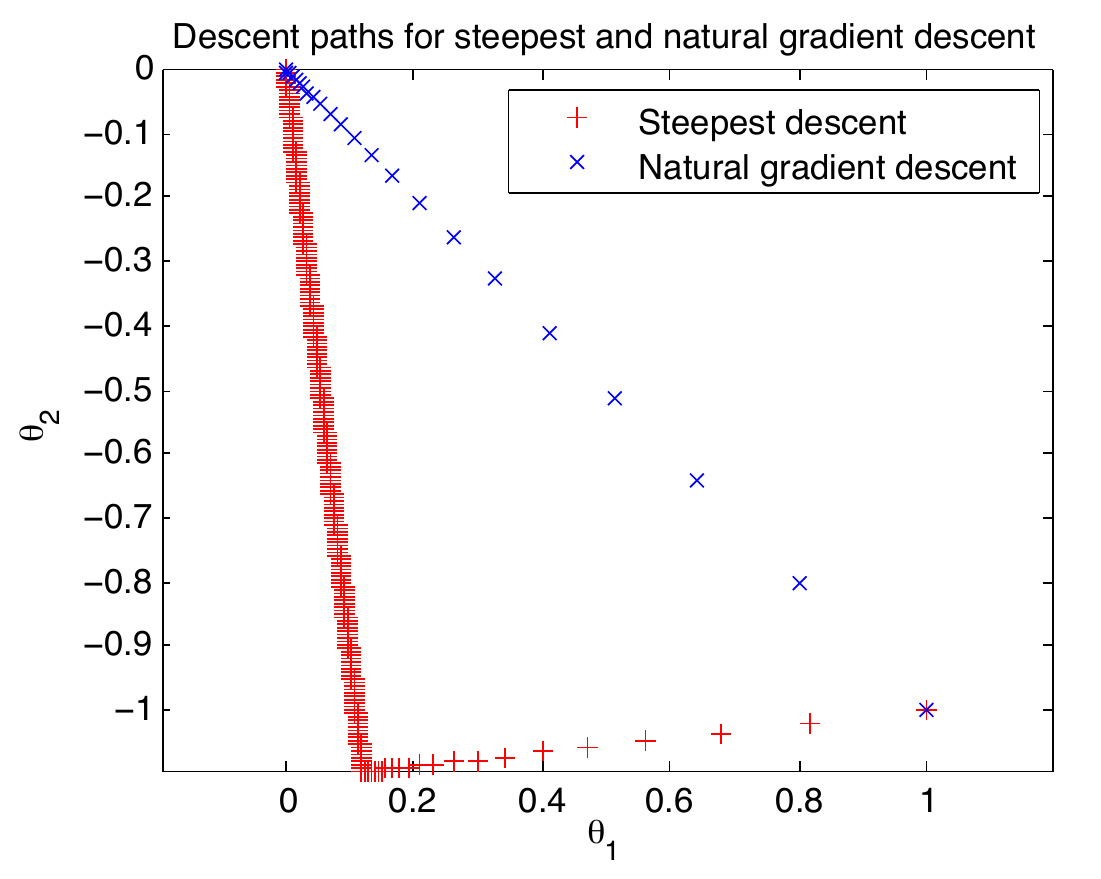}
%\begin{tabular}{c}\end{tabular}
&
(b)\includegraphics[width=0.4\linewidth]{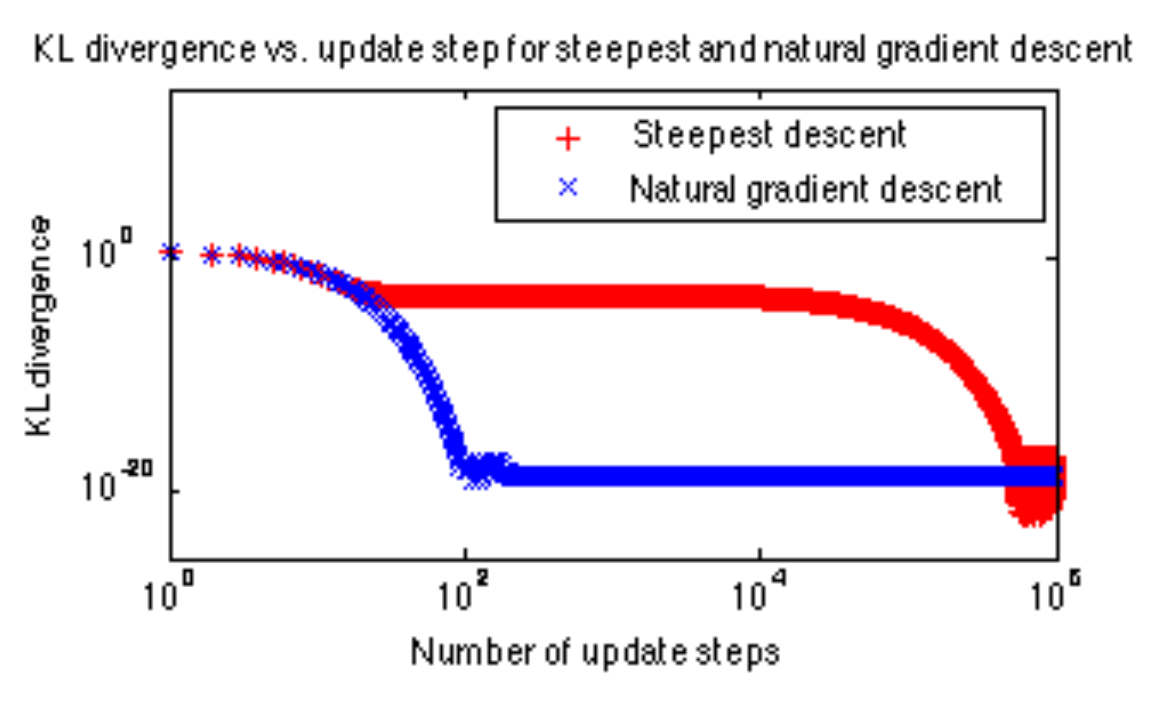}
%\begin{tabular}{c}\end{tabular}
\\
\\
(c)\includegraphics[width=0.4\linewidth]{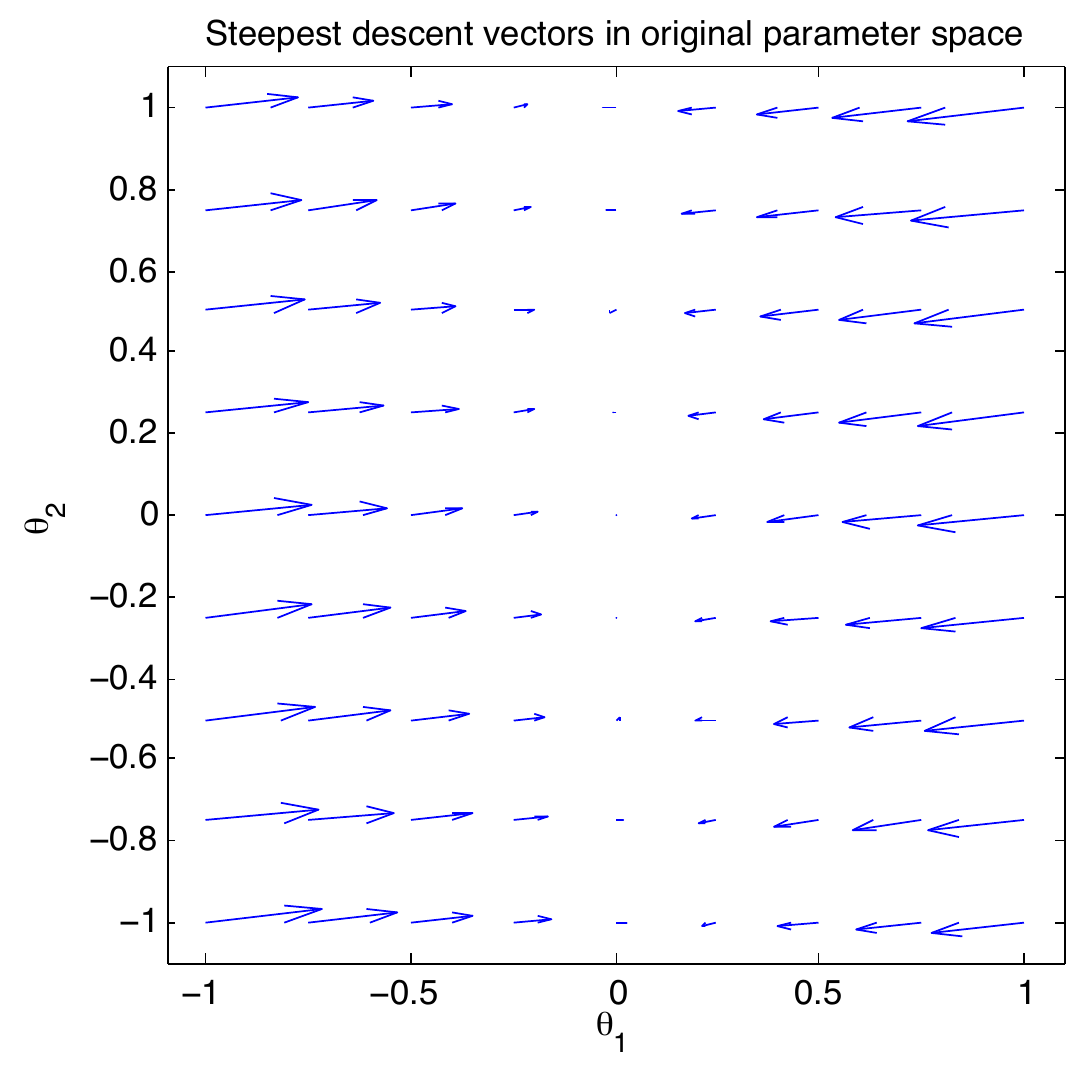}
%\begin{tabular}{c}\end{tabular}
&
(d)\includegraphics[width=0.4\linewidth]{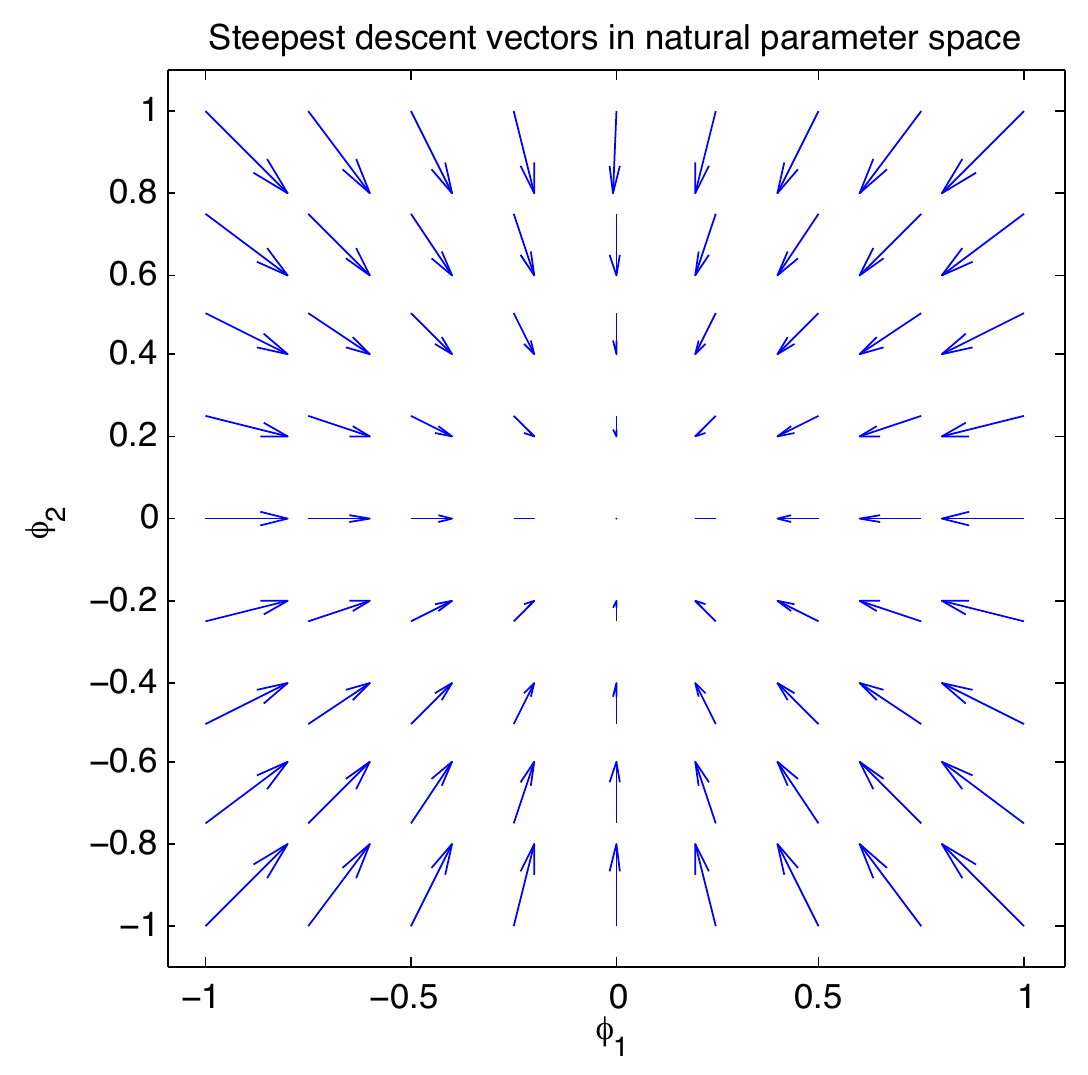}
%\begin{tabular}{c}\end{tabular}
\end{tabular}
\end{center}
\caption{
\emph{(a)} The parameter descent paths taken by steepest gradient descent (red) and natural gradient descent (blue) for the example given in Section \ref{simple example}.  The parameters are initialized at $\theta_{init} = \left[1, -1\right]^T$, and are fit to data generated with $\theta_{true} = \left[0, 0\right]^T$.  The Fisher information matrix (Equation \ref{fisher recipe}) is used to calculate the natural gradient.  Notice that steepest descent takes a more circuitous and far slower path.  
\emph{(b)} The KL divergence between the data distribution and the fit model as a function of number of gradient descent steps.  Descent using the natural gradient converges more quickly.  
\emph{(c)} The arrows give the gradient of the log likelihood objective (Equation \ref{log like obj}), for a grid of parameter settings.  This is the descent direction provided by Equation \ref{descent equation}.
\emph{(d)} The gradient of the same log likelihood objective (Equation \ref{log like obj}), but in terms of the whitened, natural, parameter space $\phi$ as described in Section \ref{whitened space}.  Note that steepest descent in the whitened space converges directly to the true parameter values $\phi_{true} = \mb G^\frac{1}{2} \theta_{true} = \left[0, 0\right]^T$.
%\emph{(a)} Distributions $p\left( \mb x \right)$ and $q\left( \mb x; \theta_1=0, \theta_2=0 \right)$ described in Section \ref{simple example} 
%\emph{(b)} Distribution $q\left( \mb x; \theta_1=0 + \Delta \theta_1, \theta_2=0 + \Delta \theta_2 \right)$ after a single steepest gradient descent step. 
%\emph{(c)} Distribution $q\left( \mb x; \theta_1=0 + {\tilde{\Delta} \theta_1}, \theta_2=0 + {\tilde{\Delta} \theta_2} \right)$ after a single natural gradient descent step. 
%\emph{(d)}Path in parameter space from initial conditions to the maximum likelihood solution, using 
%\emph{red} steepest gradient 
%and \emph{blue} natural gradient.
}
\label{gauss_pics}
\end{figure}

As can be seen in Figure \ref{gauss_pics}{\em{a}} the steepest gradient update steps can move the parameters in a direction nearly perpendicular to the desired direction.  $q\left( \mb x; \theta \right)$ is much more sensitive to changes in $\theta_1$ than $\theta_2$, so the step size in $\theta_1$ should be much smaller, but is instead much larger.  In addition, $\theta_1$ and $\theta_2$ are not independent of each other.  They move the distribution in nearly the same direction, making movement in the perpendicular direction particularly difficult.  Getting the parameters here to fully converge via steepest descent is a slow proposition, as shown in Figure \ref{gauss_pics}{\em{b}}.

The pathological learning gradient above is illustrative of a more general problem.  A model's learning gradient is effected by the parameterization of the model as well as the objective function being minimized.  The effects of the parameterization can dominate learning.  The natural gradient is a technique to remove the effects of model parameterization from learning updates.

\subsection{A metric on the parameter space}\label{sec metric}

As a first step towards compensating for differences in relative scaling, and cross-parameter dependencies, the shape of the parameter space $\theta$ is first described by assigning it a measure of distance, or a metric.  This metric is expressed via a symmetric matrix $\mb G\left( \theta \right)$, which defines the length $\left| d \theta \right|$ of an infinitesimal step $d \theta$ in the parameters,
\begin{eqnarray}
\label{metric def}
\left| d \theta \right|^2 = \sum_i \sum_j G_{ij}\left( \theta \right) d \theta_i d \theta_j = d \theta^T \mb G\left( \theta \right) d \theta
.
\end{eqnarray}
$\mb G\left( \theta \right)$ is chosen so that the length $\left| d \theta \right|$ provides a reasonable measure for the expected magnitude of the difference of $J\left( \theta + d \theta \right)$ from $J\left( \theta \right)$.  That is, $\mb G\left( \theta \right)$ is chosen such that $\left| d \theta \right|$ is representative of the expected magnitude of the change in the objective function resulting from a step $d \theta$.  There is no uniquely correct choice for $\mb G\left( \theta \right)$. %, but it must be chosen to ensure that $\left| d \theta \right|$ is invariant to transformation of the parameters $\theta$.  That is, given a new parameterization $\theta' = T\left( \theta \right)$ it must hold that $\left| d\theta \right|^2 = \left| d\theta' \right|^2$, where $d\theta' \equiv T\left( \theta + d\theta \right) - T\left( \theta \right)$.

If the objective function $J\left( \theta \right)$ is the log likelihood of a probability distribution $q\left( \mb x; \theta \right)$, then a measure of the information distance between $q\left( \mb x; \theta + d \theta \right)$ and $q\left( \mb x; \theta \right)$ usually works well, and the Fisher information matrix (Equation \ref{fisher recipe}) is frequently used as a metric.  Plugging in the example from Section \ref{simple example}, the resulting Fisher information matrix is $\mb G = \left[\begin{array}{cc}3^2 + \frac{1}{3^2} & 1 \\1 & \frac{1}{3^2}\end{array}\right]$.
%\begin{eqnarray}
%\mb G = \left< \nabla_\theta J\left( \theta \right) \left(\nabla_\theta J\left( \theta \right)\right)^T \right>_{q\left( \mb x; \theta \right)} = \left[\begin{array}{cc}3^2 + \frac{1}{3}^2 & 1 \\1 & 0\end{array}\right]
%.
%\end{eqnarray}

%\begin{figure}
%\framebox[\textwidth]{
%\parbox[c]{\textwidth}{
%\center{
%\includegraphics[width= 0.6 \linewidth]{fig_natgrad/spherical_coordinates.pdf}
%}
%}
%}
%\caption{
%A simple example of a metric.  A function $f\left( \theta_\alpha, \theta_\beta \right)$ is defined on a sphere, where $\theta_\alpha \in [-\pi, \pi)$ is an azimuthal angle and $\theta_\beta \in [-\frac{\pi}{2}, \frac{\pi}{2}]$ is an elevation angle.
%Note that at the poles, changes in $\theta_\beta$ correspond to much larger movements on the surface of the sphere than changes in $\theta_\alpha$, while at the equator small steps in either $\theta_\alpha$ or $\theta_\beta$ correspond to movements on the sphere of the same length.  To capture this uneven scaling, a metric is defined which provides an accurate measure of the distance traveled on the surface of a sphere for small changes $d\mathbf\theta = \left[ d\theta_\alpha, d\theta_\beta\right]^T$ at any $\mathbf\theta = \left[ \theta_\alpha, \theta_\beta \right]^T$.
%}
%\label{fig metric}
%\end{figure}

\begin{figure}
%\framebox[\textwidth]{
\parbox[c]{\textwidth}{
\center{
(a)\includegraphics[width= 0.4\linewidth]{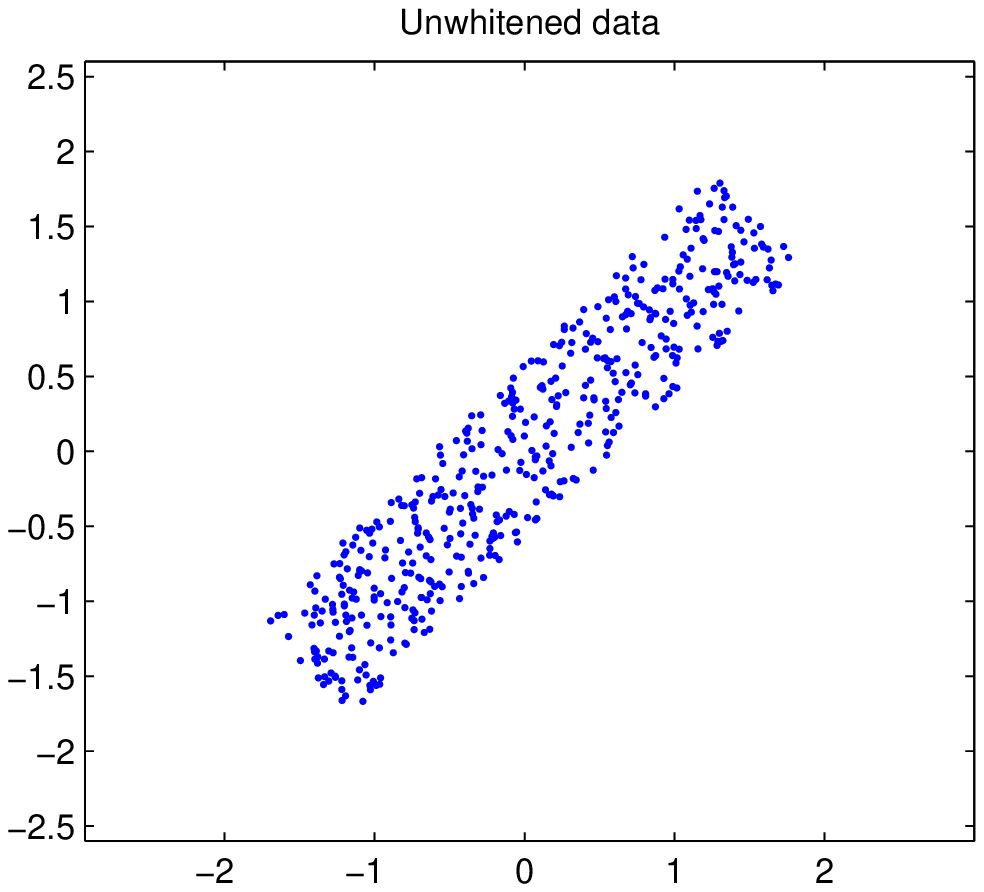} 
(b)\includegraphics[width= 0.4\linewidth]{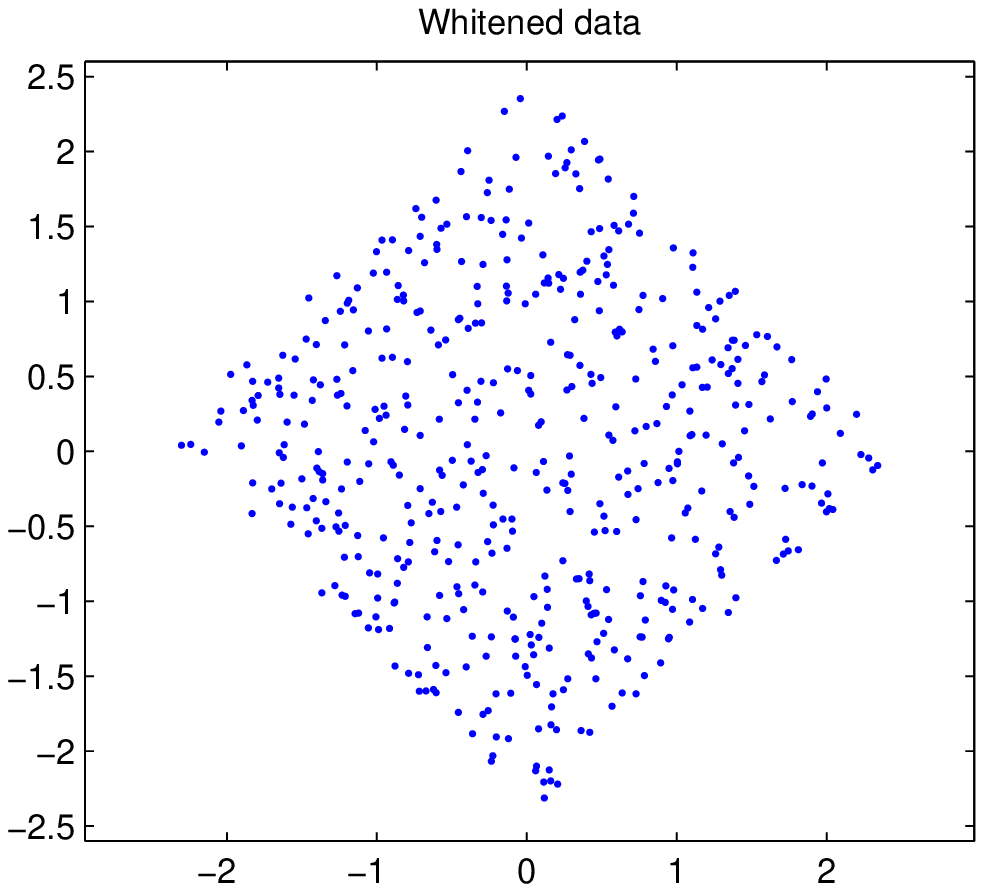}
}
}
%}
\caption{
Example of signal whitening.
\emph{(a)} Samples $\mb x$ from an unwhitened distribution in 2 variables.  
\emph{(b)} The same samples after whitening, in new variables $\mb y = \mb W\mb x = \mb\Sigma^{-\frac{1}{2}}\mb x$.%  The covariance $\mb \Sigma$ of the samples is now the identity.
}
\label{whitening example}
\end{figure}

\subsection{Connection to covariance}

$\mb G\left( \theta \right)$ is an analogue of the inverse covariance matrix $\mb\Sigma^{-1}$.  Just as a signal can be whitened given $\mb\Sigma^{-1}$ --- removing all first order dependencies and scaling the variance in each dimension to unit length --- the parameterization of $J\left( \theta \right)$ can also be ``whitened," removing the dependencies and differences in scaling between dimensions captured by $\mb G\left( \theta \right)$.  See Figure \ref{whitening example} for an example of signal whitening.  %The metric $G\left( \phi \right)$ on a whitened parameter space $\phi$ will be the identity matrix, as $\mb\Sigma^{-1}$ for a whitened signal is the identity.

As a quick review, the covariance matrix $\mb\Sigma$ of a signal $\mb x$ is defined as
\begin{eqnarray}
\mb\Sigma = \left<\mb x \mb x^T\right>
.
\end{eqnarray}
The inverse covariance matrix is frequently used as a metric on the signal $\mb x$.  This is called the Mahalanobis distance.  It has the same form as the definition of $\left| d \theta \right|^2$ in Equation \ref{metric def},
\begin{eqnarray}
\left| d\mb x \right|^2_{\mathrm{Mahalanobis}} = {d\mb x}^T \mb\Sigma^{-1} {d\mb x}
.
\end{eqnarray}

In order to whiten a signal $\mb x$, a whitening matrix $\mb W$ is found such that the covariance matrix for a new signal $\mb y = \mb W\mb x$ is the identity matrix $\mb I$.  The signal $\mb y$ is then a whitened version of $\mb x$,
\begin{eqnarray}
\mb I =  \left<\mb y \mb y^T\right> = \mb W \left<\mb x \mb x^T\right> \mb W^T = \mb W\mb\Sigma \mb W^T
.
\end{eqnarray}
Remembering that $\mb\Sigma^{-1}$ is symmetric, one solution\footnote{
Choosing $\mb W=\mb\Sigma^{-\frac{1}{2}}$ leads to symmetric, or zero-phase, whitening.  In some fields it is referred to as a decorrelation stretch.  It is equivalent to rotating a signal to the PCA basis, rescaling each axis to have unit norm, and then performing the inverse rotation, returning the signal to its original orientation.  All unitary transformations of $\mb\Sigma^{-\frac{1}{2}}$ also whiten the signal.
} to this system of linear equations is
\begin{eqnarray}
\mb W=\mb\Sigma^{-\frac{1}{2}} \\
\mb y = \mb\Sigma^{-\frac{1}{2}} \mb x
.
\end{eqnarray}
If the covariance matrix for $\mb y$ is the identity, then the metric for the Mahalanobis distance in the new variables $\mb y$ is also the identity ($\left| d\mb y \right|^2_{\mathrm{Mahalanobis}} = \mb y^T \mb y$).

Whitening is a common preprocessing step in signal processing.  It prevents incidental differences in scaling between dimensions from effecting later processing stages.

\subsection{``Whitening" the parameter space}
\label{whitened space}
If $\mb G$ is not a function of $\mb \theta$, then a similar procedure can be followed to produce a ``whitened" parameterization $\mb \phi$.  We wish to find new parameters $\mb \phi = \mb W \mb \theta$ such that the metric $\mb G$ on $\mb \phi$ is the identity $\mb I$, as the Mahalanobis metric $\mb\Sigma^{-1}$ is the identity for a whitened signal.  This will mean that a small step $d\mb  \phi$ in any direction will tend to have the same magnitude effect on the objective $J\left(\mb  \phi \right)$.%--- gradients in $\mb \phi$ will be determined by the objective function $J\left(\mb  \phi \right)$, but will not be biased by its parameterization:
\begin{align}
\phi &= \mb W \theta \\
\left| d \phi \right|^2 & = \left| d \theta \right|^2 \\
d \phi^T \mb I d \phi & = d \theta^T \mb G d \theta \\
d \phi^T d \phi & = d \theta^T \mb G d \theta \\
d \phi & = \mb W d \theta \\
d \theta^T \mb W^T \mb W d \theta & = d \theta^T \mb G d \theta
\end{align}
Noting that $\mb G$ is symmetric, we find that one solution to this system of linear equations is
\begin{eqnarray}
\mb W = \mb G^{\frac{1}{2}} \\
\phi = \mb G^{\frac{1}{2}} \theta
\end{eqnarray}
Steepest gradient descent steps in terms of $\phi$ descend the objective function in a more direct fashion than steepest gradient descent steps in terms of $\theta$, as is illustrated in Figure \ref{gauss_pics}{\em{c}} and \ref{gauss_pics}{\em{d}}.  In $\phi$, the steepest gradient is the natural gradient.

$\mb G$ is almost always a function of $\theta$, and for most problems there is no parameterization $\phi$ which will be ``white" everywhere.  So long as $\mb G\left( \theta \right)$ changes slowly though, it can be treated as constant for a single learning step.  This suggests the following as an algorithm for learning in a natural parameter space.
\begin{enumerate}
\item Express $J\left( \cdot \right)$ in terms of natural parameters $\phi = \mb G^{\frac{1}{2}}\left( \theta_t \right) \theta$.
\item Calculate an update step $\Delta \phi \propto \nabla_\phi J\left( \phi_t \right)$, where $\phi_t = \mb G^{\frac{1}{2}}\left( \theta_t \right) \theta_t$.
\item Calculate the $\theta_{t+1} = \mb G^{-\frac{1}{2}}\left( \theta_t \right) \left( \phi_t + \Delta \phi \right)$ associated with the update to $\phi$.
\item Repeat.\footnote{Practically, $\mb G\left( \theta \right)$ can usually be treated as constant for many learning steps.  This allows the natural gradient to be combined in a plug and play fashion with other gradient descent algorithms, like L-BFGS, by performing gradient descent on $J\left(\phi\right)$ rather than $J\left(\theta\right)$.}
\end{enumerate}
The resulting update steps more directly and rapidly descend the objective function than steepest descent steps.

\subsection{The natural gradient in $\theta$}

The parameter updates in Section \ref{whitened space} can be performed entirely in the original parameter space $\theta$.  
%The updates eThe procedure  If one is performing steepest gradient descent of $J\left(\phi\right)$ in the whitened parameter space $\phi$, then the equivalent update steps can be written in the original parameter space $\theta$.  
The natural gradient $\tilde{\nabla}_\theta J\left( \theta \right)$ is the direction in $\theta$ which is equivalent to steepest gradient descent in $\phi$ of $J\left( \phi \right)$. 
In order to find $\tilde{\nabla}_\theta J\left( \theta \right)$, we first write $\Delta \phi$ in terms of $\theta$, then we write the natural gradient update step in $\theta$, $\tilde{\Delta} \theta$, in terms of $\Delta \phi$, %When not combining the natural gradient with other learning techniques the change of variables can be dispensed with entirely, and the natural gradient $\tilde{\nabla}_\theta J\left( \theta \right)$ can be written directly in terms of $\mb G\left( \theta \right)$ and the true gradient $\nabla_\theta J\left( \theta \right)$.  
\begin{eqnarray}
\Delta \phi & \propto &
	 \nabla_\phi J\left( \phi \right) \\
& = &
\left(
	 \frac
		{\partial \theta}
		{\partial \phi^T}\right)^T
	\nabla_\theta J\left( \theta \right) \\
& = &
	\mb G^{-\frac{1}{2}}
	\nabla_\theta J\left( \theta \right) 
\end{eqnarray}
(where $\mathbf{\frac{\partial \theta}{\partial \phi^T}}$ is the Jacobian matrix),
\begin{eqnarray}
\tilde{\Delta} \theta & \propto &
	\frac
		{\partial \theta}
		{\partial \phi^T}
	\Delta \phi \\
& = &
	\mb G^{-\frac{1}{2}}
	\Delta \phi \\
& \propto &
	\mb G^{-1}
	\nabla_\theta J\left( \theta \right) 
.
\label{nat_grad_step}
\end{eqnarray}

Since the natural gradient update step is proportional to the natural gradient, $\tilde{\Delta} \theta \propto \tilde{\nabla}_\theta J\left( \theta \right)$, the natural gradient can be written as
\begin{eqnarray}
\tilde{\nabla}_\theta J\left( \theta \right)
= 
\mb G^{-1}\left( \theta \right)
	\nabla_\theta J\left( \theta \right)
\label{nat_grad_eq}
\end{eqnarray}

Figure \ref{gauss_pics}{\em{a}} illustrates this gradient applied to the example objective function from Section \ref{simple example}.  If gradient descent is performed by infinitesimal steps in the direction indicated by $\tilde{\nabla}_\theta J\left( \theta \right)$, then the parameterization of the problem will have no effect on the path taken during learning (though choice of $\mb G\left( \theta \right)$ will have an effect).

\section{Recipes and tricks}

In this section we present a reference with key formulas for using the natural gradient, as well as approaches useful for applying the natural gradient in specific cases. 

\subsection{Natural gradient}

The natural gradient is
\begin{eqnarray}
\tilde{\nabla}_\theta J\left( \theta \right)
=
\mb G^{-1}\left( \theta \right)
	\nabla_\theta J\left( \theta \right)
\end{eqnarray}
where $J\left( \theta \right)$ is an objective function to be minimized with parameters $\theta$, and $\mb G\left( \theta \right)$ is a metric on the parameter space.  Learning should be performed with an update rule
\begin{eqnarray}
\theta_{t+1} =  \theta_t + \tilde{\Delta} \theta_t \\
\tilde{\Delta} \theta \propto
-\tilde{\nabla}_\theta J\left( \theta \right)
\end{eqnarray}
with steps taken in the direction given by the natural gradient.

\subsection{Metric $\mb G\left( \theta \right)$}

If the objective function $J\left( \theta \right)$ is the negative log likelihood of a probabilistic model $q\left( \mb x; \theta \right)$ under an observed data distribution $p\left( \mb x \right)$
\begin{eqnarray}
J\left( \theta \right) = 
-\left< \log q\left( \mb x; \theta \right) \right>_{p\left( \mb x \right)}
\label{acceptable form}
\end{eqnarray}
then the Fisher information matrix
\begin{eqnarray}
G_{ij}\left( \theta \right) = \left< 
	\frac
		{\partial \log q\left( \mb x; \theta \right)}{\partial \theta_i}
	\frac
		{\partial \log q\left( \mb x; \theta \right)}{\partial \theta_j}
	 \right>_{q\left( \mb x; \theta \right)}
\label{fisher recipe}
\end{eqnarray}
is a good metric to use.

If the objective function is \textit{not} of of the form given in Equation \ref{acceptable form}, and cannot be transformed into that form, then greater creativity is required.  See Section \ref{non probabilistic} for some basic hints.

Remember, as will be discussed in Section \ref{sec fail}, even if the metric you choose is approximate, it is still likely to accelerate convergence!

\subsection{Fisher information over data distribution}

The Fisher information matrix (Equation \ref{fisher recipe}) requires averaging over the model distribution $q\left( \mb x; \theta \right)$.  For some models this is very difficult to do.  If that is the case, instead taking the average over the empirical data distribution $p\left( \mb x \right)$
\begin{eqnarray}
G_{ij}\left( \theta \right) = \left< 
	\frac
		{\partial \log q\left( \mb x; \theta \right)}{\partial \theta_i}
	\frac
		{\partial \log q\left( \mb x; \theta \right)}{\partial \theta_j}
	 \right>_{p\left( \mb x \right)}
\label{fisher data}
\end{eqnarray}
is frequently an effective alternative.

\subsection{Energy approximation}

Parameter estimation in a probabilistic model of the form
\begin{equation}
\label{basicmodel}
q(\mathbf{x}) = \frac{e^{-E\left(\mathbf{x};\theta\right)}}{Z\left(\theta\right)}
\end{equation}
is in general very difficult, since it requires working with the frequently intractable partition function integral $Z(\theta) = \int{e^{-E(\mathbf{x};\theta)}d\mathbf{x}}$.  There are a number of techniques which can provide approximate learning gradients (eg minimum probability flow \cite{MPF_ICML,SohlDickstein2011a}%,SohlDickstein2009p8641}
, contrastive divergence \cite{Welling:2002p3,Hinton02}, score matching \cite{Hyvarinen05}, mean field theory, and variational bayes \cite{Tanaka:1998p1984,Kappen:1997p6,Jaakkola:1997p4985,haykin2008nnc}).  Turning those gradients into natural gradients is difficult though, as the Fisher information depends on the gradient of $\log Z\left( \theta \right)$.  Practically, simply ignoring the $\log Z\left( \theta \right)$ terms entirely and using a metric
\begin{eqnarray}
G_{ij}\left( \theta \right) = \left< 
	\frac
		{\partial E\left( \mb x; \theta \right)}{\partial \theta_i}
	\frac
		{\partial E\left( \mb x; \theta \right)}{\partial \theta_j}
	 \right>_{p\left( \mb x \right)}
\label{fisher data}
\end{eqnarray}
averaged over the data distribution works surprisingly well, and frequently greatly accelerates learning.

\subsection{Diagonal approximation}

$\mb G\left( \theta \right)$ is a square matrix of size $N\times N$, where $N$ is the number of parameters in the vector $\theta$.  For problems with large $N$, $\mb G^{-1}\left( \theta \right)$ can be impractically expensive to compute and apply.  For almost all problems however, the natural gradient still improves convergence even when off-diagonal elements of $\mb G\left( \theta \right)$ are neglected,
\begin{eqnarray}
G_{ij}\left( \theta \right) = \delta_{ij} \left< \left(
	\frac
		{\partial \log q\left( \mb x; \theta \right)}{\partial \theta_i}
	\right)^2  \right>_{q\left( \mb x; \theta \right)}
,
\label{fisher diagonal}
\end{eqnarray}
making inversion and application cost $O\left( N \right)$ to perform.

If the parameters can be divided up into several distinct classes (for instance the covariance matrix and means of a gaussian distribution), block diagonal forms may also be worth considering.

\subsection{Regularization}

Even if evaluating the full $\mb G$ is easy for your problem, you may still find that $\mb G^{-1}$ is ill conditioned\footnote{This is a general problem when taking matrix inverses.  A matrix $\mb A$ with random elements, or with noisy elements, will tend to have a few very very small eigenvalues.  The eigenvalues of $\mb A^{-1}$ are the inverses of the eigenvalues of $\mb A$.  $\mb A^{-1}$ will thus tend to have a few very very large eigenvalues, which will tend to make the elements of $\mb A^{-1}$ very very large.  Even worse, the eigenvalues and eigenvectors which most dominate $\mb A^{-1}$ are those which were smallest, noisiest and least trustworthy in $\mb A$.}.  Dealing with this --- solving a set of linear equations subject to some regularization, rather than using an unstable matrix inverse --- is an entire field of study in computer science.  Here we give one simple plug and play technique, called stochastic robust approximation (Section 6.4.1 in \cite{boyd2004convex}), for regularizing the matrix inverse.  If $\mb G^{-1}$ is replaced with
\begin{equation}
\mb G^{-1}_{reg} =  \left( \mb G^T \mb G + \epsilon \I \right)^{-1}\mb G^T
\end{equation}
where $\epsilon$ is some small constant (say $0.01$), the matrix inverse will be much better behaved.

Alternatively, techniques such as ridge regression can be used to solve the linear equation
\begin{eqnarray}
\mb G\left( \theta \right) \tilde{\nabla}_\theta J\left( \theta \right) 
= 
\nabla_\theta J\left( \theta \right)
\end{eqnarray}
for $\tilde{\nabla}_\theta J\left( \theta \right)$.

\subsection{Combining the natural gradient with other techniques using the natural parameter space $\phi$ \label{nat space combine}}

It can be useful to combine the natural gradient with other gradient descent techniques.  Blindly replacing all gradients with natural gradients frequently causes problems (line search implementations, for instance, depend on the gradients they are passed being the true gradients of the function they are descending).  For a fixed value of $\mb G$ though there is a natural parameter space.
\begin{eqnarray}
\phi = \mb G^{\frac{1}{2}}\left( \theta_{fixed} \right) \theta
\end{eqnarray}
in which the steepest gradient is the same as the natural gradient.

In order to easily combine the natural gradient with other gradient descent techniques, fix $\theta_{fixed}$ to the initial value of $\theta$ and perform gradient descent over $\phi$ using any preferred algorithm.  After a significant number of update steps convert back to $\theta$, update $\theta_{fixed}$ to the new value of $\theta$, and continue gradient descent in the new $\phi$ space.

\subsection{Natural gradient of non-probabilistic models \label{non probabilistic}}

The techniques presented here are not unique to probabilistic models.  The natural gradient can be used in any context where a suitable metric can be written for the parameters.  There are several approaches to writing an appropriate metric.

\begin{enumerate}

\item If the objective function is of a form
\begin{eqnarray}
J\left( \theta \right) = \left<l\left( \mb x; \theta \right)\right>_{p(x)}
\end{eqnarray}
where $\left< \cdot \right>_{p(x)}$ indicates averaging over some data distribution $p(x)$, then it is sensible to choose a metric based on
\begin{eqnarray}
G_{ij}\left( \theta \right) & = & \left< 
	\frac
		{\partial l\left( \mb x; \theta \right)}{\partial \theta_i}
	\frac
		{\partial l\left( \mb x; \theta \right)}{\partial \theta_j}
	 \right>_{p\left( \mb x \right)}
\end{eqnarray}

\item Similarly, the penalty function can be treated as if it is the log likelihood of a probabilistic model, and the corresponding Fisher information matrix used.

For example, the task of minimizing an L2 penalty function $\left|\left| \mb y - \mb f\left(\mb x; \theta \right) \right|\right|^2$ over observed pairs of data $p\left( \mb x, \mb y \right)$ can be made probabilistic.  Imagine that the L2 penalty instead represents a conditional gaussian $q\left( \mb y | \mb x ;\theta \right) \propto \exp \left( - \left|\left| \mb y - \mb f\left(\mb x; \theta \right) \right|\right|^2 \right)$ over $\mb y$, and use the observed marginal $p\left( \mb x \right)$ over $\mb x$ to build a joint distribution $q\left( \mb x, \mb y ; \theta \right) = q\left( \mb y | \mb x ;\theta \right) p\left( \mb x \right)$.\footnote{Amari \cite{Amari1998} suggests using some uninformative model distribution $q\left( \mb x \right)$ over the inputs, such as a gaussian distribution, rather than taking $p\left( \mb x \right)$ from the data.  Either approach will likely work well.%This may be more stable, or faster to compute, but also involves more algebra and a final distribution which is farther from the observed data distribution.
}  This generates the metric:
\begin{eqnarray}
G_{ij}\left( \theta \right) & = & \left< 
	\frac
		{\partial \log \left[ q\left( \mb y|\mb x; \theta \right) p\left( \mb x \right) \right]}{\partial \theta_i}
	\frac
		{\partial \log \left[ q\left( \mb y|\mb x; \theta \right) p\left( \mb x \right) \right]}{\partial \theta_j}
	 \right>_{q\left( \mb y|\mb x; \theta \right) p\left( \mb x \right)} \\
 & = & \left< 
	\frac
		{\partial \log q\left( \mb y|\mb x; \theta \right) }{\partial \theta_i}
	\frac
		{\partial \log q\left( \mb y|\mb x; \theta \right) }{\partial \theta_j}
	 \right>_{q\left( \mb y|\mb x; \theta \right) p\left( \mb x \right)}
\end{eqnarray}

\item Find a set of parameter transformations $T\left( \theta \right)$ which you believe the distance measure $\left| d \theta \right|$ should be invariant to, and then find a metric $\mb G\left( \theta \right)$ such that this invariance holds.  That is find $\mb G\left( \theta \right)$ such that the following relationship holds for any invariant transformation $T\left( \theta \right)$,
\begin{eqnarray}
\left| \left(\theta + d\theta\right) - \theta \right|^2 & = \left| T\left(\theta + d\theta\right) - T\left( \theta \right) \right|^2
%d \theta^T \mb G\left( \theta \right) d \theta = T\left( d \theta \right)^T \mb G\left( T\left( \theta \right) \right) T\left( d \theta \right)
.
\end{eqnarray}
%where $T\left( d \theta \right) \equiv T\left( \theta + d \theta \right) - T\left( \theta \right)$.

A special case of this approach involves functions parametrized by a matrix, as presented in the next section.
\end{enumerate}

\subsection{$\mb W^T\mb W$}

As derived in \cite{Amari1998}, if a function depends on a (square, non-singular) matrix $\mb W$, it frequently aids learning a great deal to take

\begin{equation}
\tilde{\Delta} \mb W_{nat} \propto \frac
	{\partial J\left( \mb W \right) }
	{\partial \mb W} 
	\mb W^T \mb W
.
\end{equation}
The algebra leading to this rule is complex, but as discussed in the previous section it falls out of a demand that the distance measure $\left| d \mb W \right|$ be invariant to a set of transformations applied to $\mb W$.  In this case, those transformations are right multiplication by any (non-singular) matrix $\mb Y$.
\begin{eqnarray}
d \theta^T \mb G\left( \theta \right) d \theta = \left( d \theta Y \right)^T \mb G\left( \theta Y \right) \left( d \theta Y \right)
\end{eqnarray}

\subsection{What if my approximation of ${\Delta \theta}_{nat}$ is wrong?} \label{sec fail}

For any positive definite $\mb H$, movement in a direction
\begin{eqnarray}
\tilde{\Delta} \theta = \mb H \Delta \theta
\end{eqnarray}
will descend the objective function.  If the wrong $\mb H$ is used, gradient descent is performed in a suboptimal way \ldots which is the problem when steepest gradient descent is used as well.  Making an educated guess as to $\mb H$ rarely makes things worse, and frequently helps a great deal.  %Don't be scared to experiment.

\bibliographystyle{apalike}
\bibliography{note_natgrad}

\end{document}